\documentclass[10pt, a4paper]{article}

\usepackage{lrec-coling2024} % this is the new style
\usepackage{amsmath}
\usepackage{booktabs}
\title{MixRED: A Mix-lingual Relation Extraction Dataset}

\name{Lingxing Kong\textsuperscript{\rm 1}, 
Yougang Chu\textsuperscript{\rm 1},
Zheng Ma\textsuperscript{\rm 1},
Jianbing Zhang\textsuperscript{\rm 1, \rm 2, $\dagger$}
 \\ {\bf \large Liang He\textsuperscript{\rm 1}, 
Jiajun Chen\textsuperscript{\rm 1}} 
\thanks{$\dagger$ Corresponding author.}
} 
\address{
\textsuperscript{\rm 1}National Key Laboratory for Novel Software Technology, Nanjing University, China\\
\textsuperscript{\rm 2} School of Artificial Intelligence, Nanjing University, China \\
\{konglingxing, chuyg, maz, heliang\}@smail.nju.edu.cn \quad
\{zjb, chenjj\}@nju.edu.cn 
}

\abstract{
Relation extraction is a critical task in the field of natural language processing with numerous real-world applications. Existing research primarily focuses on monolingual relation extraction or cross-lingual enhancement for relation extraction. Yet, there remains a significant gap in understanding relation extraction in the mix-lingual (or code-switching) scenario, where individuals intermix contents from different languages within sentences, generating mix-lingual content. Due to the lack of a dedicated dataset, the effectiveness of existing relation extraction models in such a scenario is largely unexplored. To address this issue, we introduce a novel task of considering relation extraction in the mix-lingual scenario called MixRE and constructing the human-annotated dataset MixRED to support this task. In addition to constructing the MixRED dataset, we evaluate both state-of-the-art supervised models and large language models (LLMs) on MixRED, revealing their respective advantages and limitations in the mix-lingual scenario.  Furthermore, we delve into factors influencing model performance within the MixRE task and uncover promising directions for enhancing the performance of both supervised models and LLMs in this novel task. 
 \\ \newline \Keywords{relation extraction, code switching, dataset} }

\begin{document}

\maketitleabstract

\section{Introduction}
Relation extraction (RE) is a critical task in natural language processing, finding diverse applications across various domains. This task focuses on extracting relations between entity pairs within specific contexts, with existing research categorized into different classes. Research can be divided based on context complexity into two classes: sentence-level RE \cite{zhang2017position,riedel2010modeling,gardent2017creating}, which focuses on relations within a single sentence, and document-level RE \cite{yao2019docred,luan2018multi,cheng2021hacred}, which explores relations spanning multiple sentences. Additionally, based on the number of languages involved in the context, RE research can be divided into monolingual and multilingual categories. Monolingual RE \cite{zheng2017joint,wei2019novel,zhong2020frustratingly} concentrates on extracting relations within a single language, while multilingual RE \cite{min2017learning,ni2019neural} encompasses jointly encoding relational dependencies from contexts in multiple languages.
\begin{figure}[h]
\centering
\includegraphics[width=0.45\textwidth]{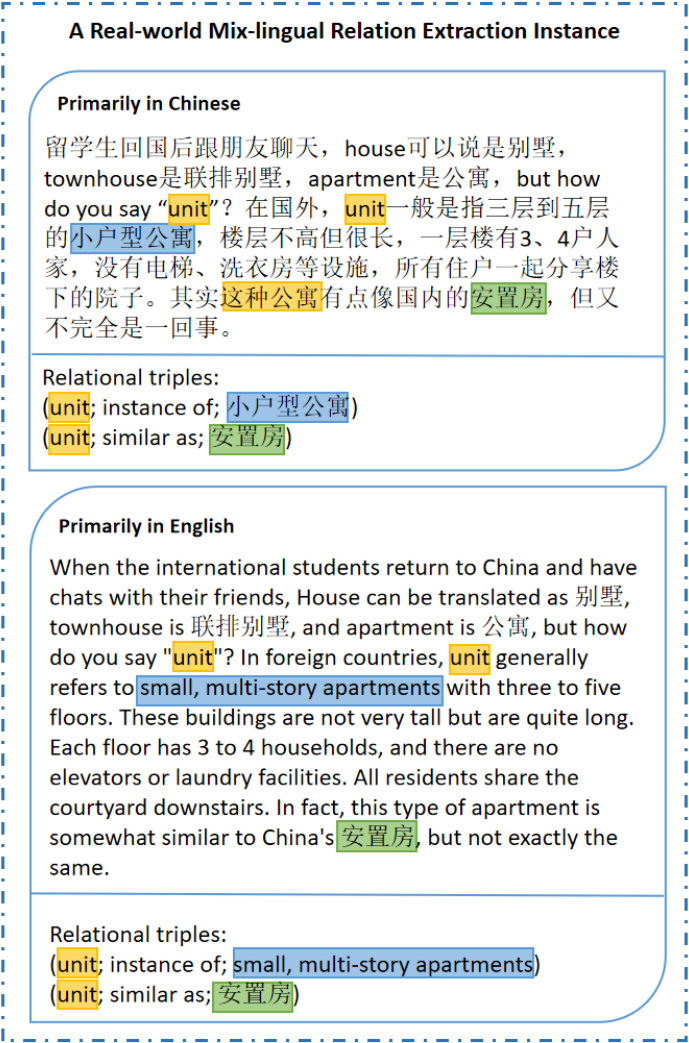} % Reduce the figure size so that it is slightly narrower than the column. Don't use precise values for figure width.This setup will avoid overfull boxes.
\caption{A real-world mix-lingual RE instance in both English and Chinese versions. Terms in the same color represent mentions of a specific entity.}
\label{Figure1}
\end{figure}

Despite the breadth of existing RE research, one crucial area remains largely unexplored: relation extraction in code-switching scenarios. Code-switching is a prevalent phenomenon where individuals blend words, phrases, or sentences from different languages to express thoughts and ideas. This practice is increasingly common in our globalized world, especially among multilingual speakers. It is observed in formal texts, such as scholarly papers comparing different cultures, as well as in informal content like social media posts by multilingual speakers. The absence of dedicated RE research in code-switching scenarios raises fundamental questions: Are existing RE models effective in such scenarios? Can multilingual models outperform monolingual ones in this context?

% To address these questions, we expand the concept of code-switching to a more intricate mix-lingual scenario and introduce a novel task called MixRE to explore relation extraction in this mix-lingual context. 
To address these issues, we consider a generalized version of the code-switching scenario named the mix-lingual scenario and introduce a novel task called MixRE to delve into relation extraction within this mix-lingual context. In contrast to prior code-switching work \cite{indra2019code}, which primarily focuses on sentence-level mixing, the mix-lingual scenario extends to document-level content. Moreover, it considers mixing at various levels, including not only contextual content but also labels, such as entities and mentions, which can significantly impact RE model performance. Figure \ref{Figure1} shows a real-world example of MixRE, showing the interweaving of sentences, phrases, mentions, and entities within a single document. In this scenario, models face the dual challenge of comprehending diverse linguistic content and capturing dependencies between entities across languages.

To support the MixRE task, we create the first human-annotated mix-lingual relation extraction dataset, MixRED, by blending English and Chinese versions of documents. In particular, we propose a systematic framework for identifying key linguistic elements within texts and substituting these elements with their semantically equivalent counterparts in another language. This process is aimed at enhancing the potential impact on model performance with the generated mix-lingual content. To adapt to the practical diversity of mix-lingual scenarios, we utilize a hierarchical mix module that employs various mixing strategies at multiple levels, including the inter-sentence level, intra-sentence level, and entity level. Furthermore, to enable a more thorough investigation into the effects of language concentration on model performance, we consider various language concentrations when generating mix-lingual samples. To ensure dataset quality, we employ well-versed bilingual Chinese-English speakers to annotate, review, and refine the constructed mix-lingual samples.

Beyond constructing the MixRED dataset, we conduct comprehensive experiments to gain insights into the novel MixRE task. We conduct an evaluation of various models, including state-of-the-art supervised models and large language models (LLMs), on the MixRED dataset to compare their performance in both mix-lingual and monolingual scenarios. Specifically, our implemented supervised models, enhanced by mix-lingual data, exhibit superior performance on MixRED, underscoring the advantage of leveraging mix-lingual data to boost model performance. Furthermore, we delve into the factors influencing model performance in the MixRE task. In detail, we explore the impact of different mix strategies and the degree of language concentration on model performance, shedding light on the potential challenges in mix-lingual contexts. Finally, we explore the utilization of mix-lingual exemplars and Chain-of-Thoughts (CoT), uncovering novel research directions for enhancing LLM performance in the mix-lingual scenario.

In summary, our contributions are threefold:

\begin{enumerate}
  \item We introduce a novel mix-lingual relation extraction task, supported by the first human-annotated dataset, MixRED.
  \item We evaluate various state-of-the-art supervised models and LLMs on MixRED and explore different factors that affect model performance in mix-lingual contexts.
  \item We identify promising directions to enhance the performance of supervised models and LLMs in the MixRE task.
\end{enumerate}
\section{Related Work}
The field of relation extraction (RE) has witnessed the development of numerous datasets, each crafted to address specific RE scenarios with unique attributes. Notably, sentence-level datasets like ACE04\footnote{\url{catalog.ldc.upenn.edu/LDC2005T09}}, ACE05\footnote{\url{catalog.ldc.upenn.edu/LDC2006T06}}, and TACRED \cite{zhang2017position}, which are enriched with entity type information, have been usually adopted to support pipeline approaches. In contrast, sentence-level datasets such as NYT \cite{riedel2010modeling} and WebNLG \cite{gardent2017creating} are commonly adopted for joint approaches. For a more comprehensive understanding of relations spanning multiple sentences, document-level datasets like DocRED \cite{yao2019docred}, SciERC \cite{luan2018multi}, and HacRED \cite{cheng2021hacred} have been introduced. These datasets emphasize the extraction of relations across extended textual contexts. Additionally, FewRel \cite{han2018fewrel} caters to the few-shot RE scenario, offering unique challenges and opportunities for model evaluation. 
In the context of multilingual RE, ACE05 includes samples in three languages: Chinese, English, and Arabic, providing valuable resources for research into sentence-level multilingual and cross-lingual RE. More recently, a bilingual document-level RE dataset named HistRED \cite{yang2023histred} has been introduced. HistRED features samples comprising historical documents in Korean and Hanja, further diversifying the landscape of multilingual RE datasets. However, it's important to note that none of the existing datasets have ventured into relation extraction within a mix-lingual scenario, leaving an intriguing research gap yet to be explored.
\section{Dataset}
\begin{figure*}[h]
\centering
\includegraphics[width=1.0\textwidth]{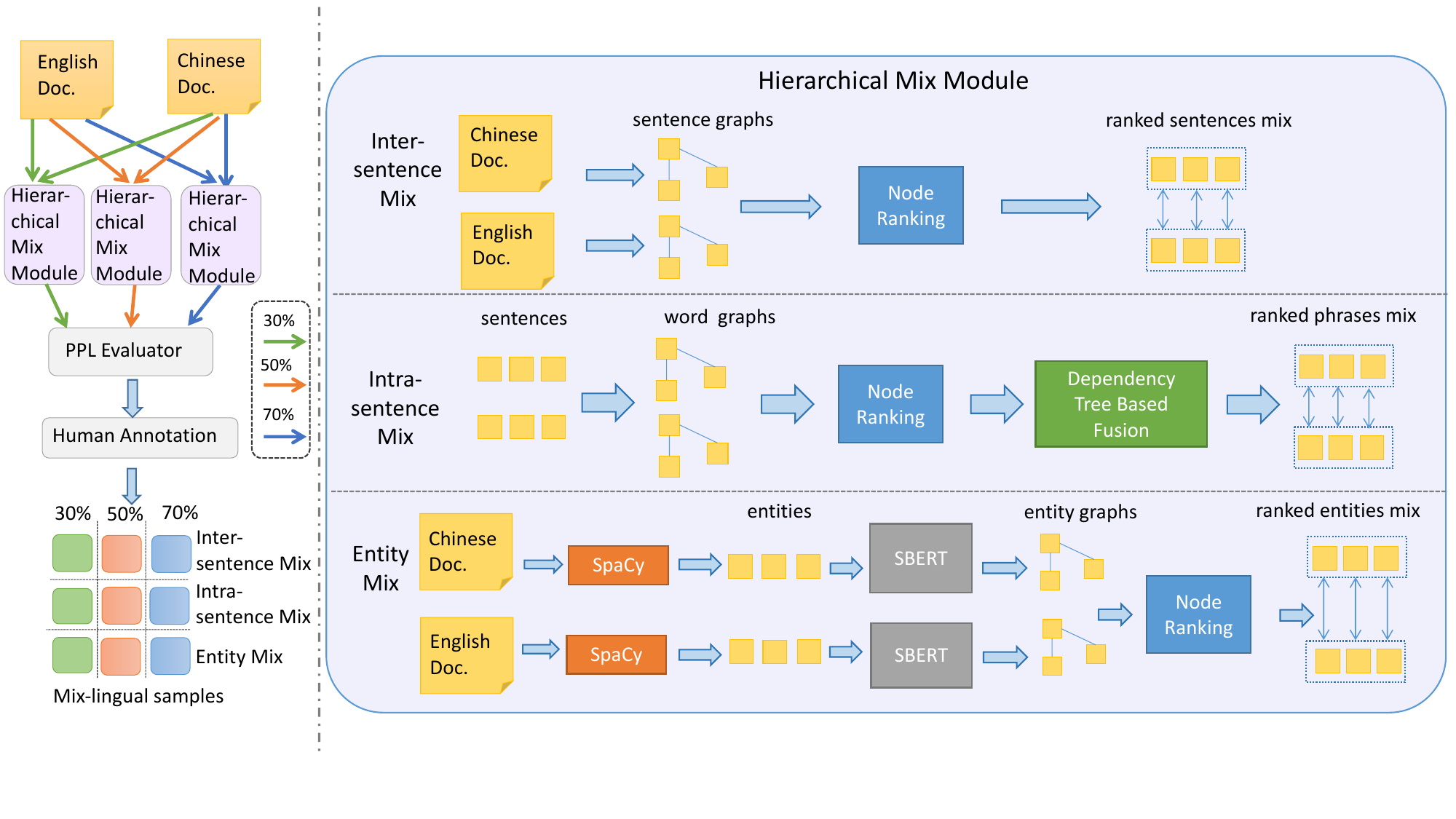} % Reduce the figure size so that it is slightly narrower than the column. Don't use precise values for figure width.This setup will avoid overfull boxes.
\caption{The construction framework of MixRED. The percentages 30\%, 50\%, and 70\% represent varying concentrations of content converted from English to Chinese during the creation of mix-lingual samples.}
\label{Figure2}
\end{figure*}
Constructing mix-lingual RE dataset samples via LLMs, such as GPT3.5, might seem intuitive. However, we observed that randomly generated mix-lingual samples by LLMs only cover singular mix-lingual or code-switching patterns, often limited to the switching of specific nouns. This falls short of encompassing the diverse mix-lingual patterns found in practice, including inter-sentence switching, intra-sentence switching, and tag switching patterns, as discussed in sociolinguistic studies of code-switching phenomena \cite{poplack1980sometimes}. Consequently, to ensure the creation of high-quality mix-lingual samples, as shown in Figure \ref{Figure2}, we developed a systematic framework. This framework is designed to construct mix-lingual samples that cover various mixing levels and different language concentrations, with a focus on crucial elements at each level, aiming to maximize their potential impact on model contextual understanding. To further enhance data quality, we engage human annotators to meticulously modify the mix-lingual sentences and create associated labels.

\subsection{Raw Documents Collection}
MixRED is constructed using raw data from an English learning website that provides VOA news reports in both English and Chinese. Several characteristics of this data source make it suitable for creating a high-quality document-level, mix-lingual RE dataset. Firstly, the news documents found in this source describe complete events within relatively lengthy documents, often exceeding 400 words. This ensures the logical coherence of the documents while simultaneously presenting challenges for RE models to capture extended dependencies between distant entities. Secondly, the news report documents encompass a wide array of topics, thereby facilitating the definition of various entity and relation types from the text. Thirdly, the English and Chinese versions of the documents are manually aligned, simplifying the blending of content from both languages.

\subsection{Mixing Different Lingual Data}
To generate mix-lingual texts, we employ a hierarchical mix module operating at three distinct levels: inter-sentence, intra-sentence, and entity level.

\subsubsection{Inter-Sentence Mix}
At the inter-sentence level, our objective is to identify sentences crucial for conveying the document's ideas. These key sentences are then replaced with their counterparts in the other language to create mix-lingual documents. We prioritize these important sentences for replacement to maximize the models' understanding of cross-sentence relational dependencies, as they often play a central role in linking other sentences. We utilize the node ranking method \cite{mihalcea2004textrank} to rank the sentences within a document in a specific language. In particular, we obtain sentence embeddings $E$ for a document from the pre-trained SBERT \cite{reimers-2019-sentence-bert}. Subsequently, we construct an undirected sentence graph for each document, with sentences as nodes and bilingual similarity scores $S$ between sentences as the weights of the edges. To calculate $S$ between sentences, we consider linguistic dimensions as follows:
$$S=\operatorname{cos\_similarity}(E_i^1,E_j^1)+\operatorname{cos\_similarity}(E_i^2,E_j^2)$$ 
Where $E_i^1$, $E_j^1$ are sentence embeddings from one language, and $E_i^2$, $E_j^2$ are from the other. With this constructed graph, the node ranking method \cite{mihalcea2004textrank} is then applied to obtain sentence rankings, where the top-ranked sentences have higher probabilities of being selected for replacement with their counterparts in the other language.

\subsubsection{Intra-Sentence Mix}
At the intra-sentence level, our aim is to identify key phrases within sentences and replace them with their equivalents in another language. To achieve this, we first obtain the word embeddings within each sentence. Subsequently, we construct a word graph based on the similarity of these word embeddings, with words serving as nodes and their similarity scores as the weights of the edges. We then employ the node ranking method \cite{mihalcea2004textrank} to identify keywords within each sentence.

Additionally, with the help of the sentence's dependency tree, we expand the selected keywords into phrases. The dependency tree reveals the internal grammatical structure of the sentence, allowing us to locate the tree nodes corresponding to the keywords and expand these nodes based on predefined patterns (e.g., adjective-noun, verb-noun). This process enables us to identify phrases that match these patterns and contain the keywords. Finally, we replace these phrases in the original sentence with their counterparts in another language.

\subsubsection{Entity Mix}
At the entity level, our goal is to replace entities within documents with their counterparts in another language. Initially, we utilize SpaCy \cite{Honnibal_spaCy_Industrial-strength_Natural_2020} to locate entities and all their mentions within the documents. When selecting entities for replacement, we pay particular attention to "entity bias." This widespread issue, identified by \citeauthor{wang2022should}, arises when existing RE models pretend to predict relations based solely on entities without considering the context of the sentence. This issue may result in incorrect predictions. For example, due to entity bias, the model might incorrectly predict the relation between "Donald Trump" and "the United States" as "President" even in a sentence like "Donald Trump once was a real estate tycoon in the United States" that does not imply such a relation.

To mitigate the influence of entity biases on the model's learning process, we take a filtering approach to filter out entities that exhibit a high degree of bias when obtaining results from SpaCy. Specifically, we employ SBERT as a zero-shot RE model, utilizing the causal inference method proposed by \citeauthor{wang2022should} to evaluate the bias degree during the RE process. 
We leverage SBERT to predict relations, using only entity pairs as input, devoid of any contextual information. If a relation can still be predicted with a high probability under these conditions, it indicates a high degree of bias towards those entity pairs, as no contextual semantics is considered. Consequently, we construct an entity bias graph, where nodes represent entities and edge weights are the predictive probabilities generated by SBERT.
Through the node ranking method \cite{mihalcea2004textrank}, the highly biased entities in the entity graph are selected and are then replaced with the other lingual counterparts to create entity-level mix-lingual samples. The replacement of the highly biased entities serves to reduce the overall bias degree of the MixRED dataset.
\subsection{Language Concentration}
During the construction of MixRED, we take into account the language concentration $L$ which is calculated as follows:
$$L = \frac{R_1}{R_2}$$
where $R_1$ represents the replaced elements converted from English to Chinese and $R_2$ represents all replaceable elements across various mixing levels in English content.
Specifically, concentrations are set at 30\%, 50\%, and 70\% when mixing the two lingual contents at each level. The inclusion of different language concentrations in MixRED facilitates the exploration of the impacts of language concentration on model performance.

To ensure document quality after mixing different lingual data and to ensure a certain concentration will not have a detrimental impact on the article's smoothness, we employ an LLM (GPT2) to calculate the perplexity (PPL) of the document mixed at a certain language concentration. We then set a threshold perplexity value to retain documents with low perplexity in the final constructed dataset.

\subsection{Human Modification and Annotation}
To ensure the quality of the dataset, we engage well-educated graduate students proficient in both Chinese and English to examine, label, and refine the constructed mix-lingual samples. To reduce the annotation workload, annotators are divided into two groups. The collected news documents are categorized into topics based on their content. Each topic is assigned to a specific annotator group for the initial annotation. In the first pass, annotators label the mentions, entities, and relations within each document sample. They also assess the smoothness of the mix-lingual contents to ensure quality. After the first pass, annotator groups rotate topics for the second pass. This process involves further scrutiny of labels, sentence syntax, and sentence semantics to ensure accuracy and consistency. The labeled samples are then combined, disregarding the original topics, and randomly distributed among the annotators for the third pass. This approach ensures that each annotator evaluates samples across various topics, enhancing the overall quality and diversity of annotations. To assess consistency between different passes and among annotators, we use the weighted Cohen's Kappa as a metric. We set a threshold value of 0.8 to identify batches of samples with questionable agreement. Any disagreements are documented and subjected to further examination

\subsection{Dataset Analysis}
We present a comparative analysis of MixRED alongside existing RE datasets, as summarized in Table \ref{table1}. MixRED exhibits exceptional characteristics across various dimensions, including mix-lingual samples, relation distribution, and data complexity.
\begin{table*}
\centering
\begin{tabular}{lccccccc} 
\toprule
                              & \textbf{\#Doc.} & \textbf{\#Sent.} & \textbf{\#Word} & \textbf{\#Ent.} & \textbf{\#Mention} & \textbf{\#Rel.} & \textbf{Avg. Mention}  \\ 
\midrule
\textbf{SemEval 2010 Task8}             & -              & 10,717          & 205k            & 21,434          & 21,434             & 9               & 1                      \\
\textbf{ACE2003\textasciitilde{}2004} & -              & 12,783          & 297k            & 46,108          & 46,108             & 24              & 1                      \\
\textbf{TACRED}                        & -              & 53,791          & 1,823k          & 152,527         & 152,527            & 41              & 1                      \\
\textbf{FewRel}                        & -              & 56,109          & 1,397k          & 72,124          & 72,124             & 100             & 1                      \\
\textbf{DocRED}                        & 5,053          & 40,276          & 1,002k          & 98,560          & 128,128            & 96              & 1.3                    \\
\textbf{MixRED}                  & 8,187          & 237,952         & 3,639k          & 44,043          & 91,176             & 21              & 2.1                    \\
\bottomrule
\end{tabular}
\caption{Comparison of MixRED with existing RE datasets.}
\label{table1}
\end{table*}

\begin{figure}[h]
\centering
\includegraphics[width=0.45\textwidth]{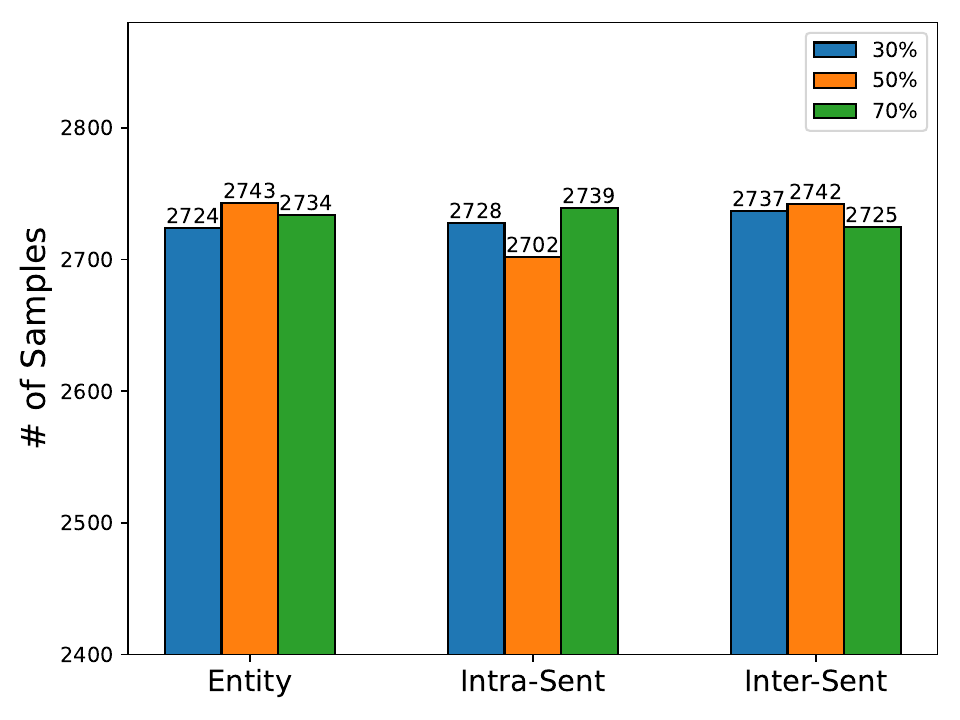} % Reduce the figure size so that it is slightly narrower than the column. Don't use precise values for figure width.This setup will avoid overfull boxes.
\caption{Distribution of samples in MixRED.}
\label{Figure3}
\end{figure}
\subsubsection{Mix-lingual Samples}
MixRED stands as the first document-level RE dataset to encompass both the Chinese and English languages. Given that English and Chinese are two of the most widely spoken languages globally, the exploration of their synergistic use holds substantial practical relevance. Beyond providing mix-lingual documents, MixRED also offers monolingual versions in both Chinese and English, facilitating bilingual research. Importantly, during the creation of mix-lingual samples, we maintain a balanced distribution of samples to ensure equitable evaluation of model performance in the MixRE task. As shown in Figure \ref{Figure3}, MixRED effectively maintains both an even distribution of samples across different mix levels and an even distribution of samples across varying concentrations within each mix level.
\subsubsection{Relation Distribution}
\begin{figure}[h]
\centering
\includegraphics[width=0.5\textwidth]{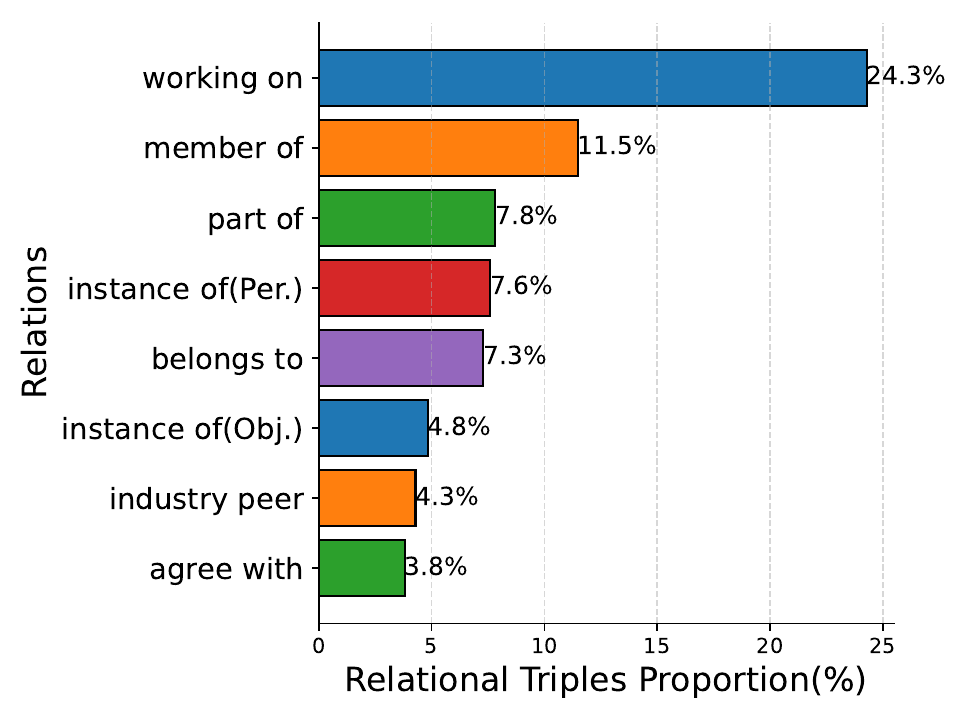} % Reduce the figure size so that it is slightly narrower than the column. Don't use precise values for figure width.This setup will avoid overfull boxes.
\caption{Distribution of relational triples for the top 40\% relations in MixRED.}
\label{Figure4}
\end{figure}

MixRED effectively addresses the issue of uneven relation distribution that has plagued previous datasets. This issue pertains to the concentration of relational triples in existing datasets around a few biased relation types. For instance, in NYT, TACRED,
and WebNLG, the top 20\% relations overwhelmingly dominate 98.93\%, 91.33\%, and 77.57\% of total relational triples, respectively \cite{cheng2021hacred}. In contrast, MixRED significantly mitigates this bias, with the top 20\% relations representing only 50.02\% of the total, and even the top 40\% relations represent only 70.64\% of the total. Figure \ref{Figure4} provides a detailed view of the distribution of relational triples for the top 40\% relations in MixRED, demonstrating a balanced tripe distribution over most relation types.
\subsubsection{Data Complexity}
Table \ref{table1} exemplifies MixRED's exceptional data complexity, surpassing other datasets in multiple aspects. MixRED boasts the highest count of total sentences and words, with an average document length of 444 words, which exceeds that of established datasets such as DocRED (198 words). This abundance of content and longer dependencies within MixRED introduces additional challenges for RE models, demanding effective relation extraction across sentences. Furthermore, MixRED includes a wealth of entity mentions in both languages, with the average number of entity mentions exceeding that of any other dataset. This abundance of mentions not only underscores the dataset's complexity but also provides a valuable resource for studying the impacts of multilingual mentions on model performance.

\section{Experiment}
\subsection{Evaluated Models}
In our evaluation, we consider both supervised models and large language models (LLMs) on the MixRED dataset. The supervised models include monolingual document-level RE models: \textbf{LSR} \cite{nan2020reasoning}, \textbf{BERT-E} \cite{zhou2021document}, and \textbf{ATLOP} \cite{zhou2021document}. 
Another evaluated model is \textbf{BERT-E-xlmr} within which we replace the encoder of BERT-E from BERT to XLM-R \cite{conneau2020unsupervised}, a language model pre-trained in multilingual scenarios. Furthermore, based on KMLM \cite{liu2022enhancing}, which utilizes code-switching data for knowledge-oriented pre-training, we implement two mix-lingual RE models: \textbf{ATLOP-mix} and \textbf{BERT-E-mix}, by replacing the context encoder of ATLOP and BERT-E with KMLM. For the LLMs, we evaluate \textbf{GPT3.5}\footnote{https://openai.com/blog/chatgpt/}, \textbf{BLOOMZ} \cite{muennighoff2022crosslingual}, \textbf{Baichuan} \cite{yang2023baichuan}, \textbf{LLaMA2-7B} \cite{cui2023efficient}, and \textbf{LLaMA2-13B} \cite{cui2023efficient}.
\subsection{Setup}
In our experiments, we partition MixRED\footnote{We plan to make the dataset available at: https://github.com/acddca/MixRED} into training, development, and testing sets, distributed in a 6:2:2 ratio. We tune hyperparameters for all models on the development set. The supervised models are adapted to MixRED by modifying input and output formats as necessary. We employ the F1 score as our evaluation metric and report results averaged over three runs for each experiment. In our experiments with LLMs, we employ a one-round chat strategy. This involves providing documents containing entity pairs to the models, allowing them to generate predictions for all potential relations within the documents.

\subsection{Model Performance}

\begin{table*}
\centering
    \begin{tabular}{lcccc} 
    \toprule
                            & \textbf{DocRED} & \textbf{MixRED-English} & \textbf{MixRED-Chinese} & \textbf{MixRED}  \\ 
    \midrule
    \textbf{Supervised Models}  & \textbf{F1}     & \textbf{F1}             & \textbf{F1}             & \textbf{F1}      \\
    LSR                         & 59.0            & 32.9                    & 31.6                    & 31.0             \\
    BERT-E                      & 56.3            & 35.6                    & 36.9                    & 32.4             \\
    ATLOP                       & 61.5            & 35.4                    & 39.5                    & 31.3             \\
    \textbf{Enhanced Models} &                 &                         &                         &                  \\
    BERT-E-XLM-R                       & -               & 36.1                    & 31.4                    & 36.2             \\
    BERT-E-mix                  & -               & 37.6                    & 34.0                    & 37.1             \\
    ATLOP-mix                   & -               & 38.0                    & 37.5                    & 37.4             \\
    \textbf{LLMs}               &                 &                         &                         &                  \\
    GPT3.5                     & -               & 11.4                    & 17.8                    & 13.3             \\
    Baichuan                    & -               & 4.6                     & 5.8                     & 3.6              \\
    BLOOMZ                      & -               & 1.7                     & 0.1                     & 0.1              \\
    LLaMA2-7B                    & -               & 7.1                     & 2.5                     & 4.9              \\
    LLaMA2-13B                   & -               & 7.8                     & 9.0                     & 8.2              \\
    \bottomrule
    \end{tabular}

\caption{Evaluation of the SOTA supervised models and LLMs on DocRED, monolingual versions of MixRED, and MixRED.}
\label{table2}
\end{table*}
Table \ref{table2} provides a comparative analysis of model performance on Docred, MixRED, and the monolingual versions of MixRED (MixRED-Chinese and MixRED-English). This comparison reveals intriguing insights into model capabilities in the MixRE task. When evaluating supervised models on MixRED, we observe a substantial performance decrease compared to Docred. Furthermore, their performance on the monolingual versions of MixRED also exhibits a significant decline relative to Docred. This decline emphasizes that MixRED's multilingual nature and its complex content, which includes multiple mentions, introduce unique challenges for existing models. 

When comparing supervised models on MixRED-Chinese, MixRED-English, and MixRED, we find that the models' performance is consistently lower on MixRED. This pattern underscores that the mix-lingual scenario is inherently more challenging for existing supervised models than monolingual scenarios. Notably, our implemented mix-lingual model ATLOP-mix, outperforms the best monolingual model (BERT-E) by 5.0 F1 scores and surpasses the multilingual model (BERT-E-XLM-R) by 1.2 F1 scores on MixRED. This outcome signifies that models can indeed benefit from mix-lingual data during pretraining, with the mix-lingual patterns learned by models significantly enhancing their understanding of relational dependencies in mix-lingual contexts. Moreover, the mix-lingual models' performance on MixRED closely approaches their performance on the monolingual versions of MixRED, further underscoring the suitability of these models for mix-lingual scenarios.

Comparing the performance of supervised models and LLMs, we find that supervised models consistently outperform LLMs in the MixRE task. This aligns with previous reports \cite{li2023evaluating} highlighting that supervised models also outperform LLMs in traditional sentence-level RE tasks. This disparity underscores that, despite LLMs' substantial achievements in various NLP tasks, a significant gap still exists between LLMs and supervised models in understanding relational dependencies across diverse settings. Interestingly, we observe diverse behaviors among LLMs on MixRED and its monolingual versions. While models like Baichuan and BLOOMZ experience a significant performance decline on MixRED compared to the monolingual versions of MixRED, other models like GPT3.5, LLaMA2-7B, and LLaMA2-13B exhibit a performance level on MixRED that falls between their performance on the monolingual versions of MixRED. This diversity in LLM behavior can be attributed to the varying proportions of multilingual corpora employed by different LLMs during their pretraining, indicating the need for further research to ascertain the most effective combinations and proportions of multilingual corpora to enhance LLM adaptation to mix-lingual scenarios.

\subsection{Factors Analysis}
\begin{table}
\centering
\resizebox{\linewidth}{!}{
    \begin{tabular}{lccc} 
    \toprule
                    & \textbf{Inter-sentence~} & \textbf{Intra-sentence} & \textbf{Entity}  \\ 
    \midrule
    \textbf{Models} & \textbf{F1}              & \textbf{F1}             & \textbf{F1}      \\
    LSR             & 33.3                     & 33.8                    & 27.5             \\
    BERT-E          & 34.3                     & 34.7                    & 29.3             \\
    ATLOP           & 34.5                     & 34.1                    & 30.8             \\
    GPT3.5         & 11.3                     & 11.0                    & 11.8             \\
    LLaMA2-7B        & 7.1                      & 5.7                     & 4.8              \\
    LLaMA2-13B       & 8.0                      & 7.4                     & 6.7              \\
    \bottomrule
    \end{tabular}
}
\caption{Model performance across different mix levels.}
\label{table3}
\end{table}

\begin{table}
\centering
\resizebox{\linewidth}{!}{
    \begin{tabular}{lccc} 
    \toprule
                    & \textbf{MixRED-30\%} & \textbf{MixRED-50\%} & \textbf{MixRED-70\%}  \\ 
    \midrule
    \textbf{Models} & \textbf{F1}          & \textbf{F1}          & \textbf{F1}           \\
    LSR             & 33.7                 & 32.0                 & 31.9                  \\
    BERT-E          & 36.8                 & 36.5                 & 36.1                  \\
    ATLOP           & 36.5                 & 35.8                 & 35.0                  \\
    GPT3.5         & 12.1                 & 13.3                 & 12.2                  \\
    LLaMA2-7B        & 6.3                  & 5.1                  & 5.2                   \\
    LLaMA2-13B       & 7.2                  & 7.9                  & 8.1                   \\
    \bottomrule
    \end{tabular}
}
\caption{Model performance across different language concentrations. The percentages 30\%, 50\%, and 70\% represent varying concentrations of content converted from English to Chinese during the creation of mix-lingual samples.}
\label{table4}
\end{table}
To gain a deeper understanding of the factors influencing model performance in the MixRE task, we conduct additional experiments, exploring the impacts of different mix levels and language concentrations.
\subsubsection{Impact of Mix Levels} 
We partition MixRED into subsets of different mix levels and assess model performance on these subsets, as shown in Table \ref{table3}. We notice for all supervised models, performance remains similar between the inter-sentence mix and intra-sentence mix subsets but significantly drops in the entity mix subset. This suggests that it is challenging for supervised models to capture dependencies between mix-lingual entities. Furthermore, it underscores the sensitivity of supervised RE models to entity variations, as also discussed in \cite{wang2022should}. Intriguingly, LLMs do not exhibit a substantial performance decline in the entity mix subset. In fact, the best-performing LLM, GPT3.5, performs even better on the entity mix subset than on the other subsets. This observation implies that LLMs are less sensitive to entity variations than supervised models in the MixRE task.
\subsubsection{Impact of Language Concentration} 
We segment MixRED into subsets with varying language concentrations and evaluate model performance on these subsets, as shown in Table \ref{table4}. The results demonstrate that for supervised models, performance consistently deteriorates as the concentration of Chinese contents increases. This trend is expected since these models are primarily based on language models pretrained with English corpus, rendering them less adept at handling scenarios rich in content from other languages. However, this trend does not hold for LLMs. Instead, we notice different LLMs exhibit proficiency at different language concentrations. We deduce that this variation arises from the diverse proportions of multilingual corpora used during the pretraining of LLMs, which in turn results in varying preferences for language concentrations among these models.

\subsection{Directions for Enhancing LLM performance}
\begin{figure}[h]
\centering
\includegraphics[width=0.5\textwidth]{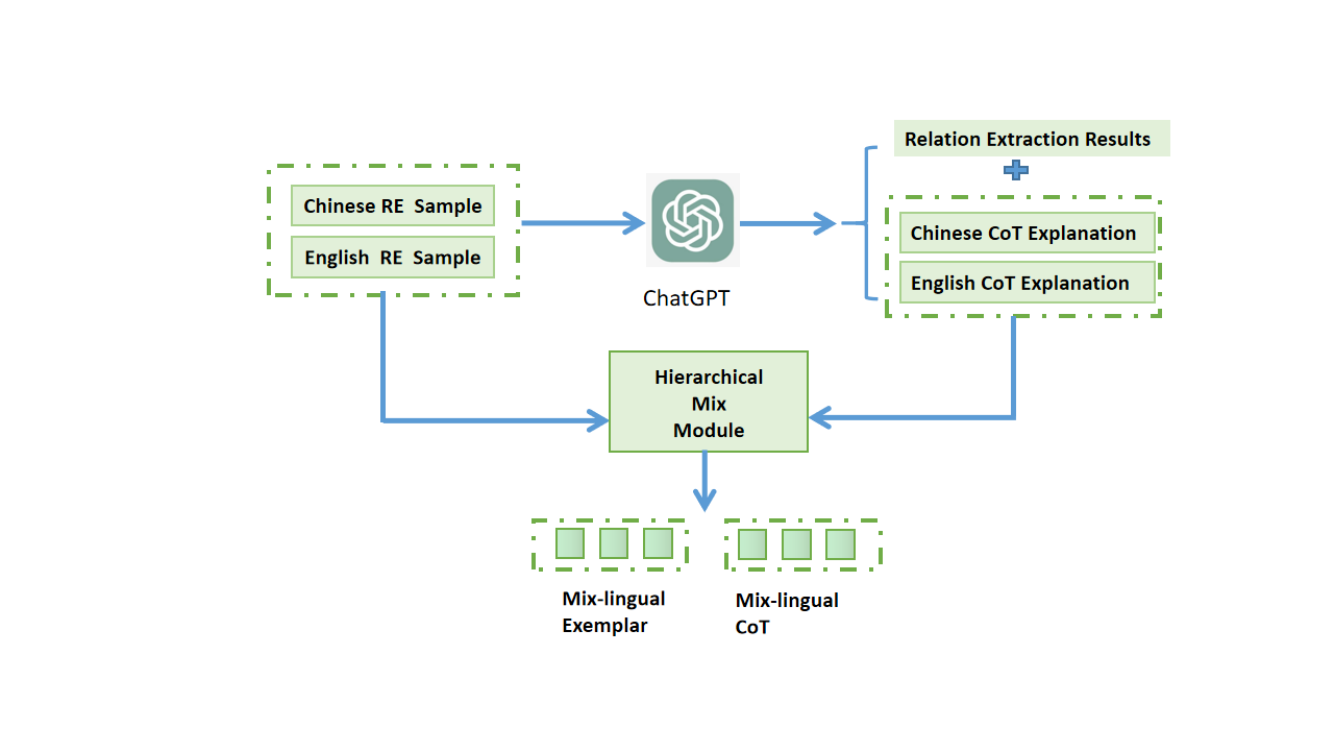} % Reduce the figure size so that it is slightly narrower than the column. Don't use precise values for figure width.This setup will avoid overfull boxes.
\caption{Development of mix-lingual exemplars and CoT for enhancing LLM performance.}
\label{Figure5}
\end{figure}
We dive deeper into the enhancements that can boost the performance of LLMs in the MixRE task. Specifically, we investigate the impact of different combinations of exemplars and Chain-of-Thoughts (CoT) when used in the prompts provided to the LLMs. Regarding exemplars, we move beyond monolingual RE exemplars and introduce the concept of mix-lingual exemplars. These mix-lingual exemplars are designed to assist LLMs in grasping the intricacies of mix-lingual patterns. In the context of CoT, we take inspiration from previous work \cite{wadhwa2023revisiting} and employ GPT-3.5 to generate few-shot CoT explanations, which are then included in the prompts for all LLMs in our experiments. These CoT explanations also comprise both monolingual and mix-lingual versions. Crucially, the creation of mix-lingual exemplars and CoT explanations follows the same hierarchical mix module used in constructing MixRED samples. This approach ensures a consistent representation of the diverse mix-lingual patterns within the generated data. The construction process for mix-lingual exemplars and CoT is shown in Figure \ref{Figure5}.
\begin{figure}[h]
\centering
\includegraphics[width=0.5\textwidth]{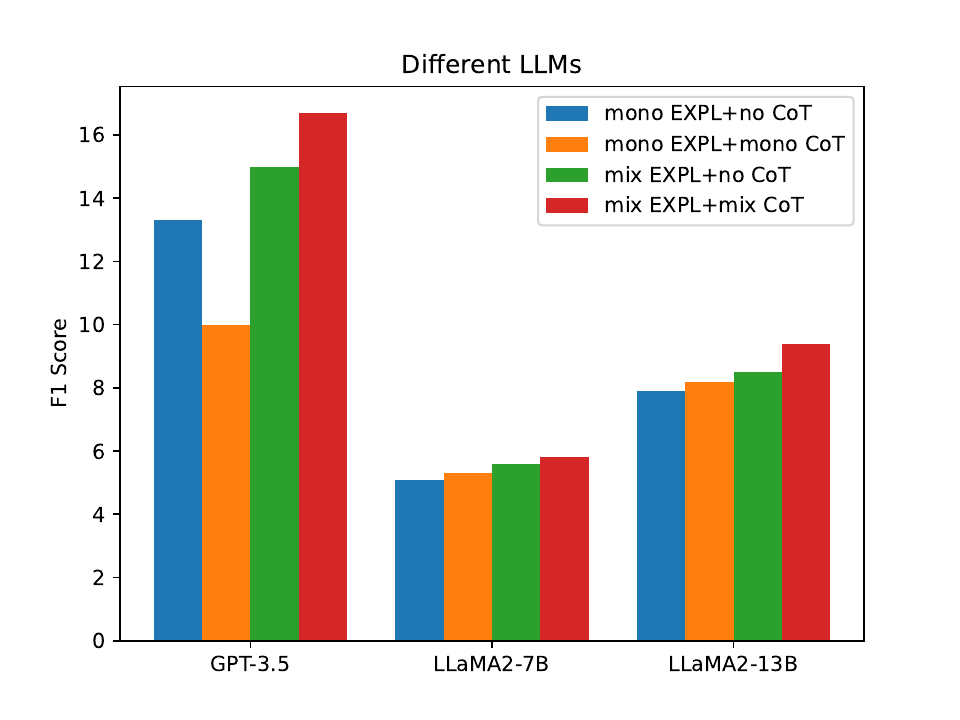} % Reduce the figure size so that it is slightly narrower than the column. Don't use precise values for figure width.This setup will avoid overfull boxes.
\caption{Comparative evaluation of LLM performance with diverse exemplar and CoT combinations. In the figure, "mono" designates monolingual, "mix" signifies mix-lingual, and "EXPL" stands for exemplar.}
\label{Figure6}
\end{figure}

A comparative analysis of the impacts of various exemplar and CoT combinations across different LLMs is detailed in Figure \ref{Figure6}. We notice that when exemplars are utilized in isolation, the inclusion of mix-lingual exemplars leads to substantial performance enhancements across all LLMs, surpassing the improvements gained by using monolingual exemplars alone. Furthermore, the introduction of mix-lingual CoT further augments LLM performance when compared to using mix-lingual exemplars exclusively. Notably, the combination of mix-lingual exemplars and mix-lingual CoT outperforms all other combinations in terms of performance gains. These results underscore the contributions of both mix-lingual exemplars and CoT to enhancing LLMs' capacity to comprehend relational dependencies in the MixRE task. These findings serve as an inspirational guide for the development of novel instructional approaches for LLMs in future research endeavors.

\section{Conclusion}
In this study, we have introduced a novel task, MixRE, designed for relation extraction in mix-lingual contexts. To enable this novel task, we have meticulously constructed MixRED, the very first human-annotated mix-lingual RE dataset. The construction of MixRED hinges on a systematic framework wherein we leverage a hierarchical mix strategy and take into account various language concentration degrees, thus yielding diverse mix-lingual samples. Also, we conduct extensive experiments on MixRED. Initially, we compare the performance of a spectrum of supervised models and LLMs on this dataset. Subsequently, we delved into the influence of different mix strategies and language concentrations on model performance. Furthermore, we undertook a thorough examination of LLM behaviors in the MixRE task, unearthing valuable insights for improving LLM performance.

\section{References}
\label{lr:ref}
\bibliographystyle{lrec-coling2024-natbib}
\bibliography{lrec-coling2024-example}

\begin{thebibliography}{28}
\expandafter\ifx\csname natexlab\endcsname\relax\def\natexlab#1{#1}\fi

\bibitem[{Cheng et~al.(2021)Cheng, Liu, Qu, Zhao, Liang, Wang, Huai, Yuan, and
  Xiao}]{cheng2021hacred}
Qiao Cheng, Juntao Liu, Xiaoye Qu, Jin Zhao, Jiaqing Liang, Zhefeng Wang,
  Baoxing Huai, Nicholas~Jing Yuan, and Yanghua Xiao. 2021.
\newblock Hacred: A large-scale relation extraction dataset toward hard cases
  in practical applications.
\newblock In \emph{Findings of the Association for Computational Linguistics:
  ACL-IJCNLP 2021}, pages 2819--2831.

\bibitem[{Conneau et~al.(2020)Conneau, Khandelwal, Goyal, Chaudhary, Wenzek,
  Guzm{\'a}n, Grave, Ott, Zettlemoyer, and Stoyanov}]{conneau2020unsupervised}
Alexis Conneau, Kartikay Khandelwal, Naman Goyal, Vishrav Chaudhary, Guillaume
  Wenzek, Francisco Guzm{\'a}n, {\'E}douard Grave, Myle Ott, Luke Zettlemoyer,
  and Veselin Stoyanov. 2020.
\newblock Unsupervised cross-lingual representation learning at scale.
\newblock In \emph{Proceedings of the 58th Annual Meeting of the Association
  for Computational Linguistics}, pages 8440--8451.

\bibitem[{Cui et~al.(2023)Cui, Yang, and Yao}]{cui2023efficient}
Yiming Cui, Ziqing Yang, and Xin Yao. 2023.
\newblock Efficient and effective text encoding for chinese llama and alpaca.
\newblock \emph{arXiv preprint arXiv:2304.08177}.

\bibitem[{Gardent et~al.(2017)Gardent, Shimorina, Narayan, and
  Perez-Beltrachini}]{gardent2017creating}
Claire Gardent, Anastasia Shimorina, Shashi Narayan, and Laura
  Perez-Beltrachini. 2017.
\newblock Creating training corpora for nlg micro-planning.
\newblock In \emph{55th annual meeting of the Association for Computational
  Linguistics (ACL)}.

\bibitem[{Han et~al.(2018)Han, Zhu, Yu, Wang, Yao, Liu, and
  Sun}]{han2018fewrel}
Xu~Han, Hao Zhu, Pengfei Yu, Ziyun Wang, Yuan Yao, Zhiyuan Liu, and Maosong
  Sun. 2018.
\newblock Fewrel: A large-scale supervised few-shot relation classification
  dataset with state-of-the-art evaluation.
\newblock \emph{arXiv preprint arXiv:1810.10147}.

\bibitem[{Honnibal et~al.(2020)Honnibal, Montani, Van~Landeghem, and
  Boyd}]{Honnibal_spaCy_Industrial-strength_Natural_2020}
Matthew Honnibal, Ines Montani, Sofie Van~Landeghem, and Adriane Boyd. 2020.
\newblock \href {https://doi.org/10.5281/zenodo.1212303} {{spaCy:
  Industrial-strength Natural Language Processing in Python}}.

\bibitem[{Indra~Winata et~al.(2019)Indra~Winata, Madotto, Wu, and
  Fung}]{indra2019code}
Genta Indra~Winata, Andrea Madotto, Chien-Sheng Wu, and Pascale Fung. 2019.
\newblock Code-switched language models using neural based synthetic data from
  parallel sentences.
\newblock \emph{arXiv e-prints}, pages arXiv--1909.

\bibitem[{Li et~al.(2023)Li, Fang, Yang, Wang, Ye, Zhao, and
  Zhang}]{li2023evaluating}
Bo~Li, Gexiang Fang, Yang Yang, Quansen Wang, Wei Ye, Wen Zhao, and Shikun
  Zhang. 2023.
\newblock \href {http://arxiv.org/abs/2304.11633} {Evaluating chatgpt's
  information extraction capabilities: An assessment of performance,
  explainability, calibration, and faithfulness}.

\bibitem[{Liu et~al.(2022)Liu, Li, He, Bing, Joty, and Si}]{liu2022enhancing}
Linlin Liu, Xin Li, Ruidan He, Lidong Bing, Shafiq Joty, and Luo Si. 2022.
\newblock Enhancing multilingual language model with massive multilingual
  knowledge triples.
\newblock In \emph{Proceedings of the 2022 Conference on Empirical Methods in
  Natural Language Processing}, pages 6878--6890.

\bibitem[{Luan et~al.(2018)Luan, He, Ostendorf, and Hajishirzi}]{luan2018multi}
Yi~Luan, Luheng He, Mari Ostendorf, and Hannaneh Hajishirzi. 2018.
\newblock Multi-task identification of entities, relations, and coreference for
  scientific knowledge graph construction.
\newblock In \emph{Proceedings of the 2018 Conference on Empirical Methods in
  Natural Language Processing}. Association for Computational Linguistics.

\bibitem[{Mihalcea and Tarau(2004)}]{mihalcea2004textrank}
Rada Mihalcea and Paul Tarau. 2004.
\newblock Textrank: Bringing order into text.
\newblock In \emph{Proceedings of the 2004 conference on empirical methods in
  natural language processing}, pages 404--411.

\bibitem[{Min et~al.(2017)Min, Jiang, Freedman, and
  Weischedel}]{min2017learning}
Bonan Min, Zhuolin Jiang, Marjorie Freedman, and Ralph Weischedel. 2017.
\newblock Learning transferable representation for bilingual relation
  extraction via convolutional neural networks.
\newblock In \emph{Proceedings of the Eighth International Joint Conference on
  Natural Language Processing (Volume 1: Long Papers)}, pages 674--684.

\bibitem[{Muennighoff et~al.(2022)Muennighoff, Wang, Sutawika, Roberts,
  Biderman, Le~Scao, Saiful~Bari, Shen, Yong, Schoelkopf
  et~al.}]{muennighoff2022crosslingual}
Niklas Muennighoff, Thomas Wang, Lintang Sutawika, Adam Roberts, Stella
  Biderman, Teven Le~Scao, M~Saiful~Bari, Sheng Shen, Zheng-Xin Yong, Hailey
  Schoelkopf, et~al. 2022.
\newblock Crosslingual generalization through multitask finetuning.
\newblock \emph{arXiv e-prints}, pages arXiv--2211.

\bibitem[{Nan et~al.(2020)Nan, Guo, Sekuli{\'c}, and Lu}]{nan2020reasoning}
Guoshun Nan, Zhijiang Guo, Ivan Sekuli{\'c}, and Wei Lu. 2020.
\newblock Reasoning with latent structure refinement for document-level
  relation extraction.
\newblock In \emph{Proceedings of the 58th Annual Meeting of the Association
  for Computational Linguistics}, pages 1546--1557.

\bibitem[{Ni and Florian(2019)}]{ni2019neural}
Jian Ni and Radu Florian. 2019.
\newblock Neural cross-lingual relation extraction based on bilingual word
  embedding mapping.
\newblock In \emph{Conference on Empirical Methods in Natural Language
  Processing and International Joint Conference on Natural Language
  Processing}. Association for Computational Linguistics.

\bibitem[{Poplack(1980)}]{poplack1980sometimes}
Shana Poplack. 1980.
\newblock Sometimes i’ll start a sentence in spanish y termino en espanol:
  toward a typology of code-switching1.

\bibitem[{Reimers and Gurevych(2019)}]{reimers-2019-sentence-bert}
Nils Reimers and Iryna Gurevych. 2019.
\newblock \href {https://arxiv.org/abs/1908.10084} {Sentence-bert: Sentence
  embeddings using siamese bert-networks}.
\newblock In \emph{Proceedings of the 2019 Conference on Empirical Methods in
  Natural Language Processing}. Association for Computational Linguistics.

\bibitem[{Riedel et~al.(2010)Riedel, Yao, and McCallum}]{riedel2010modeling}
Sebastian Riedel, Limin Yao, and Andrew McCallum. 2010.
\newblock Modeling relations and their mentions without labeled text.
\newblock In \emph{Machine Learning and Knowledge Discovery in Databases:
  European Conference, ECML PKDD 2010, Barcelona, Spain, September 20-24, 2010,
  Proceedings, Part III 21}, pages 148--163. Springer.

\bibitem[{Wadhwa et~al.(2023)Wadhwa, Amir, and Wallace}]{wadhwa2023revisiting}
Somin Wadhwa, Silvio Amir, and Byron~C Wallace. 2023.
\newblock Revisiting relation extraction in the era of large language models.
\newblock \emph{arXiv e-prints}, pages arXiv--2305.

\bibitem[{Wang et~al.(2022)Wang, Chen, Zhou, Cai, Liang, Liu, Yang, Liu, and
  Hooi}]{wang2022should}
Yiwei Wang, Muhao Chen, Wenxuan Zhou, Yujun Cai, Yuxuan Liang, Dayiheng Liu,
  Baosong Yang, Juncheng Liu, and Bryan Hooi. 2022.
\newblock Should we rely on entity mentions for relation extraction? debiasing
  relation extraction with counterfactual analysis.
\newblock In \emph{Proceedings of the 2022 Conference of the North American
  Chapter of the Association for Computational Linguistics: Human Language
  Technologies}.

\bibitem[{Wei et~al.(2019)Wei, Su, Wang, Tian, and Chang}]{wei2019novel}
Zhepei Wei, Jianlin Su, Yue Wang, Yuan Tian, and Yi~Chang. 2019.
\newblock A novel cascade binary tagging framework for relational triple
  extraction.
\newblock \emph{arXiv preprint arXiv:1909.03227}.

\bibitem[{Yang et~al.(2023{\natexlab{a}})Yang, Xiao, Wang, Zhang, Bian, Yin,
  Lv, Pan, Wang, Yan, Yang, Deng, Wang, Liu, Ai, Dong, Zhao, Xu, Sun, Zhang,
  Liu, Ji, Xie, Dai, Fang, Su, Song, Liu, Ru, Ma, Wang, Liu, Lin, Nie, Guo,
  Sun, Zhang, Li, Li, Cheng, Chen, Zeng, Wang, Chen, Men, Yu, Pan, Shen, Wang,
  Li, Jiang, Gao, Zhang, Zhou, and Wu}]{yang2023baichuan}
Aiyuan Yang, Bin Xiao, Bingning Wang, Borong Zhang, Ce~Bian, Chao Yin, Chenxu
  Lv, Da~Pan, Dian Wang, Dong Yan, Fan Yang, Fei Deng, Feng Wang, Feng Liu,
  Guangwei Ai, Guosheng Dong, Haizhou Zhao, Hang Xu, Haoze Sun, Hongda Zhang,
  Hui Liu, Jiaming Ji, Jian Xie, JunTao Dai, Kun Fang, Lei Su, Liang Song,
  Lifeng Liu, Liyun Ru, Luyao Ma, Mang Wang, Mickel Liu, MingAn Lin, Nuolan
  Nie, Peidong Guo, Ruiyang Sun, Tao Zhang, Tianpeng Li, Tianyu Li, Wei Cheng,
  Weipeng Chen, Xiangrong Zeng, Xiaochuan Wang, Xiaoxi Chen, Xin Men, Xin Yu,
  Xuehai Pan, Yanjun Shen, Yiding Wang, Yiyu Li, Youxin Jiang, Yuchen Gao,
  Yupeng Zhang, Zenan Zhou, and Zhiying Wu. 2023{\natexlab{a}}.
\newblock \href {http://arxiv.org/abs/2309.10305} {Baichuan 2: Open large-scale
  language models}.

\bibitem[{Yang et~al.(2023{\natexlab{b}})Yang, Choi, Cho, and
  Choo}]{yang2023histred}
Soyoung Yang, Minseok Choi, Youngwoo Cho, and Jaegul Choo. 2023{\natexlab{b}}.
\newblock Histred: A historical document-level relation extraction dataset.
\newblock \emph{arXiv preprint arXiv:2307.04285}.

\bibitem[{Yao et~al.(2019)Yao, Ye, Li, Han, Lin, Liu, Liu, Huang, Zhou, and
  Sun}]{yao2019docred}
Yuan Yao, Deming Ye, Peng Li, Xu~Han, Yankai Lin, Zhenghao Liu, Zhiyuan Liu,
  Lixin Huang, Jie Zhou, and Maosong Sun. 2019.
\newblock Docred: A large-scale document-level relation extraction dataset.
\newblock In \emph{Proceedings of the 57th Annual Meeting of the Association
  for Computational Linguistics}, pages 764--777.

\bibitem[{Zhang et~al.(2017)Zhang, Zhong, Chen, Angeli, and
  Manning}]{zhang2017position}
Yuhao Zhang, Victor Zhong, Danqi Chen, Gabor Angeli, and Christopher~D Manning.
  2017.
\newblock Position-aware attention and supervised data improve slot filling.
\newblock In \emph{Conference on Empirical Methods in Natural Language
  Processing}.

\bibitem[{Zheng et~al.(2017)Zheng, Wang, Bao, Hao, Zhou, and
  Xu}]{zheng2017joint}
Suncong Zheng, Feng Wang, Hongyun Bao, Yuexing Hao, Peng Zhou, and Bo~Xu. 2017.
\newblock Joint extraction of entities and relations based on a novel tagging
  scheme.
\newblock \emph{arXiv preprint arXiv:1706.05075}.

\bibitem[{Zhong and Chen(2020)}]{zhong2020frustratingly}
Zexuan Zhong and Danqi Chen. 2020.
\newblock A frustratingly easy approach for entity and relation extraction.
\newblock \emph{arXiv preprint arXiv:2010.12812}.

\bibitem[{Zhou et~al.(2021)Zhou, Huang, Ma, and Huang}]{zhou2021document}
Wenxuan Zhou, Kevin Huang, Tengyu Ma, and Jing Huang. 2021.
\newblock Document-level relation extraction with adaptive thresholding and
  localized context pooling.
\newblock In \emph{Proceedings of the AAAI conference on artificial
  intelligence}, volume~35, pages 14612--14620.

\end{thebibliography}
\bibliographystylelanguageresource{lrec-coling2024-natbib}
\bibliographylanguageresource{languageresource}

\end{document}